\documentclass{article}

\usepackage{microtype}
\usepackage{graphicx}
\usepackage{subcaption}
\usepackage{multicol}

\usepackage{booktabs} 

\usepackage{hyperref}

\newcommand{\secref}[1]{\S\ref{#1}}

\usepackage[accepted]{sysml2019}

\sysmltitlerunning{PoET-BiN}

\begin{document}

\twocolumn[
\sysmltitle{PoET-BiN: Power Efficient Tiny Binary Neurons}




\begin{sysmlauthorlist}
\sysmlauthor{Sivakumar Chidambaram}{po1}
\sysmlauthor{J.M. Pierre Langlois}{po2}
\sysmlauthor{Jean Pierre David}{po1}

\end{sysmlauthorlist}

\sysmlaffiliation{po1}{Department of Electrical Engineering, Polytechnique Montreal, Montreal, Canada}
\sysmlaffiliation{po2}{Department of Computer and Software Engineering, Polytechnique Montreal, Montreal, Canada}

\sysmlcorrespondingauthor{Sivakumar Chidambaram }{csivanitw@gmail.com}

\sysmlkeywords{Machine Learning, SysML}

\vskip 0.3in

\begin{abstract}
The success of neural networks in image classification has inspired various hardware implementations on embedded platforms such as Field Programmable Gate Arrays, embedded processors and Graphical Processing Units. These embedded platforms are constrained in terms of power, which is mainly consumed by the Multiply Accumulate operations and the memory accesses for weight fetching. Quantization and pruning have been proposed to address this issue. Though effective, these techniques do not take into account the underlying architecture of the embedded hardware. In this work, we propose PoET-BiN, a Look-Up Table based power efficient implementation on resource constrained embedded devices. A modified Decision Tree approach forms the backbone of the proposed implementation in the binary domain. A LUT access consumes far less power than the equivalent Multiply Accumulate operation it replaces, and the modified Decision Tree algorithm eliminates the need for memory accesses. We applied the PoET-BiN architecture to implement the classification layers of networks trained on MNIST, SVHN and CIFAR-10 datasets, with near state-of-the art results. The energy reduction for the classifier portion reaches up to six orders of magnitude compared to a floating point implementations and up to three orders of magnitude when compared to recent binary quantized neural networks.
\end{abstract}
]



\printAffiliationsAndNotice{} 

\section{Introduction}
Neural networks form the backbone of current technologies such as face recognition \cite{A_Yang_2015_ICCV}, text comprehension \cite{B_P16-1086} and speech emulation \cite{C_Pascual2017SEGANSE}. The recent successes of neural networks can be attributed to the algorithmic advances made possible in part due to the availability of powerful computational devices and of large datasets to train the neural networks. GPUs are the mainstream devices to train them. However, most embedded applications do not require powerful GPUs, given that only the inference task needs to be implemented with pre-trained networks. Therefore, hardware accelerators built on low power devices such as FPGAs, embedded GPUs and embedded processors are preferred to efficiently run these neural networks in real-time. \cite{lacey2016deep}
\par
Implementing neural networks on embedded devices poses a set of challenges such as power consumption, latency and adaptability. A vanilla neural network can be designed to achieve the best test set accuracy without much considerations for power consumption and memory requirement. These networks use 32-bit floating point representations that require expensive MAC operations and memory read operations. Methods such as quantization of weights and activations address these challenges by reducing the data precision down to 4 bits \cite{Zhuang_2018_CVPR} for the Imagenet dataset and sometimes to the extreme extent of single bit binary values for smaller datasets such as CIFAR-10 and SVHN \cite{courbariaux2016binarized}. Surprisingly, quantization may not be detrimental to the the accuracy of the network as it can provide a form of regularization similar to dropout \cite{6639346}, which helps to better generalize on the testset. Similarly, pruning weights reduces memory reads and MAC operations while maintaining the accuracy. Such methods can actually lower the computation effort but they are still founded on the simulation of a neural network with a target architecture that is not specifically adapted to the target algorithms. \par
Field Programmable Gate Arrays (FPGA) are intrinsically closer to the architecture of a neural network because they embed millions of small computing elements, such as Look-Up Tables (LUT), that can be interconnected to form a large network of computing elements. Nevertheless, their inputs are binary signals while neural networks usually handle higher precision data. Binary Neural Networks \cite{courbariaux2016binarized} open a way to very efficient FPGA implementations. In such networks, binary multiplications are implemented with simple XNOR gates while the overall sum (accumulation) is computed using a popcount \cite{sun2016} and a final comparison of this MAC operation produces the binary output activation. However, each neuron may have a large number of inputs (1000 in the case of ResNet implementations \cite{he2016deep}) while FPGA computing resources can handle just a few inputs per LUT. Modern FPGAs have 4-6 input bits and 1-2 output bits per LUT. This mismatch leads to inefficient architecture spanning multiple LUTs for each operation when implemented on a FPGA.
\par
In this work, we build networks of tiny binary neurons and map them to FPGA LUTs, thereby providing an optimized way to implement equivalent MAC operations while achieving near state of the art accuracy. We name this architecture and its associated building algorithm {\it PoET-BiN}. It is very power efficient since most of the processing is done as Look-Up operations in the binary domain and does not require floating point multipliers, adders or external memory accesses. 
Our LUT-based architecture combines Decision Trees (DTs) and weighted sums of binary classifiers. A major motivation to use LUT-based DTs is because it eliminates memory reads and can be implemented using simple logic gates, thus considerably saving power. On the other hand, DTs alone are weak classifiers. Hence, we modify the inherently weak classifiers to solve complex non-linear classification tasks consuming a fraction of the power compared to other implementations. The following four contributions are proposed to combine DTs and weighted sums of binary signals to best exploit embedded hardware resources in the context of neural network implementations :
\begin{itemize} 
    
    \item A modified DT training algorithm to better handle a fixed number of inputs LUTs.
    \item The Reduced Input Neural Circuit (RINC) : A LUT-based architecture founded on modified DTs and the hierarchical version of the well known Adaboost algorithm to efficiently implement a network of binary neurons.
    \item A sparsely connected output layer for multiclass classification.
    \item The PoET-BiN architecture consisting of multiple RINC modules and a sparsely connected output layer.
    \item Automatic VHDL code generation of the PoET-BiN architecture for FPGA implementation.
\end{itemize}


Section 2 details the proposed PoET-BiN architecture, followed by the experimental setup in Section 3. The results and comparison with state-of-the-art methods are detailed in section 4. Finally, the related works and conclusions are presented in Section 5 and 6 respectively.

\section{The PoET-BiN Architecture} \label{ov_so}

In this section we detail the principal contribution, multi-level RINC architecture for binary feature representation (\secref{m_rinc}) and also the sparsely connected output layer for multiclass classification (\secref{bin_mul}), who together constitute PoET-BiN. 

\subsection{Multi-level RINC Architecture} \label{m_rinc}

The RINC is a network of tiny binary neurons with limited inputs. In a traditional neural network, each neuron can have up to 4096 neurons as seen in VGG Networks \cite{simonyan2014very}. This impedes efficient hardware implementation as it leads to numerous and interdependent logical circuits that adversely affect the power consumption, speed and area of the architecture. In our architecture, the number of single-bit inputs to each neuron is limited, usually less than 8. This poses a major challenge to choose the best inputs among the ones available and classifying the data based on only these selected inputs. \par
 We follow an approach inspired by DTs to implement a binary neuron ( \secref{rinc0}). The size of the DTs is limited by the number of LUT inputs. DTs are inherently weak classifiers and boosting techniques are used to group weak classifiers thus forming stronger classifiers. The Adaboost algorithm is one of the most widely used boosting algorithm. In section \ref{rinc1}, we detail our LUT-based implementation of the Adaboost algorithm. Still, the LUT-based algorithm is limited by the number of weak classifiers that can be grouped together. To further enhance our classifiers, we introduce a hierarchical Adaboost algorithm, where the number of DTs increases exponentially with every level. At each level, all the operations are designed to exactly fit in a single LUT, thereby optimizing power and area efficiency. The hierarchical algorithm is detailed in section \ref{rinc2}.\par

\subsubsection{RINC-0 : Modified Decision Tree Algorithm} \label{rinc0}
\begin{figure}[h]
      \centering
    \begin{subfigure}[t]{0.4\textwidth}
        \centering
        \includegraphics[scale=0.7]{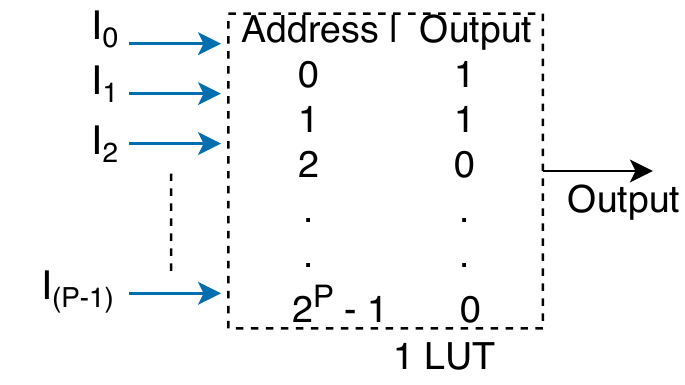}
        \caption{LUT-based binary neuron implementation}
    \end{subfigure}
    \\
    \centering
    \begin{subfigure}[t]{0.5\textwidth}
        \centering
        \includegraphics[scale=0.43]{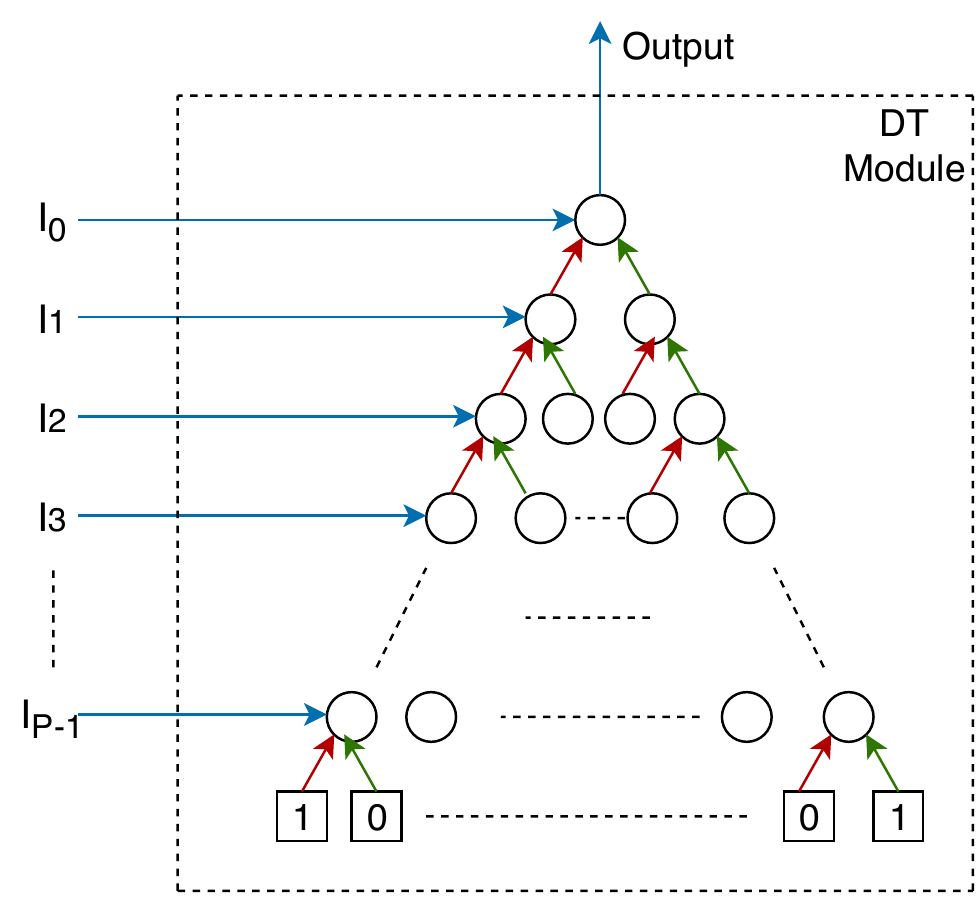}
        \caption{RINC-0 architecture with DT}
    \end{subfigure}
    \caption{LUT and its equivalent DT}
    \label{fig6}
  
  \end{figure}

    A binary neuron only has two possible outputs, so all possible input combinations can be classified into two groups. Hence, each binary neuron is a binary classifier that can be implemented as an Input vs Output table for all the possible input combinations in LUTs (Fig.\ref{fig6}). Hardcoding such tables in a LUT or a memory is feasible when the number of inputs is limited. With each added input the complexity increases by a factor of $2$. In most cases, we need to model a binary neuron with more inputs than can be accommodated in a single LUT. Therefore, hardcoding the input-output relation is not a viable option. This demands an algorithm to choose the best inputs from the input set so as to fit the implementation in a single LUT. We use a greedy approach inspired by DTs to choose the best inputs from the available set of inputs. There are many other traditional machine learning classifier such as Support Vector Machines \cite{Burges1998} and Naive Bayes classifier \cite{Langley:1992:ABC:1867135.1867170}. However, DTs provide a distinctive edge as they can be easily and efficiently implemented in LUTs \cite{abdelsalam2018polybinn}. \par
    
    The original DT algorithm \cite{Quinlan1986} is limited either by the depth or by the number of nodes, which often leads to under utilization of LUTs, because the LUTs used to implement the DTs are neither constrained by the depth or nodes, rather by the number of distinct inputs. Hence, we propose a modified DT algorithm that attempts to optimize DTs for a given number of inputs. Off-the-shelf DT classifiers are built one node at a time. Each node is associated with a feature that divides the input feature space and minimizes the entropy of the DT. Contrarily, we train DTs layer-wise. Hence, all nodes in the same level of the DT have the same features. This divides the input feature space into $2^{P}$ (where $P$ is the number of inputs to a LUT) sub-spaces and increases the capacity of the DT for a given number of input features. Moreover, the leaf nodes, which contain the possible output (0 or 1) for each combination, can be selected in $O(1)$ time. The resulting LUT-based DT is named as Reduced Input Neural Circuit (RINC-0), where-"0" signifies the level, which is further explained in the following sections. With this modified DT approach we have an increased capacity RINC-0 architecture that can fit exactly into one LUT and is limited by the number of input features.\par

    \begin{algorithm}[tb]
   \caption{RINC0: Level wise DT training algorithm}
   \label{alg:example1}
    \begin{algorithmic}
   \STATE {\bfseries Input:} data $X$, size $n \times F$
   \STATE Initialize $Used\_features  = []$
   \STATE Initialize $Label\_array = []$
   \FOR{$i \rightarrow 1$ {\bfseries to} $p$}
        \FOR{$feat \rightarrow 1$ {\bfseries to} $F$ }
            \IF{feat\ $not\ in\ $ Used\_features}
                \STATE level\_entropy $\rightarrow$ 0
                \FOR{$node \rightarrow 1$ {\bfseries to} $2^{p-1}$ }
                    \STATE Calculate\ entropy\ of\ the\ current\ node
                    
                    \STATE level\_entropy += node\_entropy
                \ENDFOR
                \IF{level\_entropy $\leq$ min\_entropy}
                    \STATE min\_entropy $\rightarrow$ level\_entropy
                    \STATE best\_feature $\rightarrow$ feat
                \ENDIF
            \ENDIF
        \ENDFOR
        \STATE Append\ Used\_features\ array\ with\ best\_feat
    \ENDFOR
    \FOR{$cur\_node \rightarrow 1$ {\bfseries to} $2^{P}$ }
        \STATE $S_0$ $\rightarrow$ Sum\ of\ $class_0$\ training\ examples\ at cur\_node  
        \STATE $S_1$ $\rightarrow$ Sum\ of\ $class_1$\ training\ examples\ at cur\_node
        \IF{$S_0 \leq S_1$}
            \STATE Append\ Label\_array with $1$
        \ELSE
            \STATE Append\ Label\_array with $0$
        \ENDIF
    \ENDFOR
    \STATE Return Label\_array, Used\_features
   
\end{algorithmic}
\end{algorithm} 
    Algorithm \ref{alg:example1} details the training algorithm for the RINC-0 module. The training dataset contains $n$ examples of $F$ dimensions each. Our goal is to choose the best $P$ features from the available $F$ features that best classify the data, or, in other words, reduce the entropy. We train a level-based DT approach where we choose the best feature ($best\_feature$) from all features that have not been used before, that reduces the entropy of the entire level to the largest extent. The best feature is appended to the Used feature array($Used\_features$). The label array ($Label\_array$) contains the class label for each leaf node. A leaf node is assigned a class that has the highest number of training examples that end up in that leaf node. These class labels form the output column of the LUT and the best features form the input indices to the LUT.\par

     Fig. \ref{fig6} illustrates a DT and its equivalent LUT. The red and green arrows at each node represent the path taken when the feature at the node is 0 or 1, respectively. This implementation of a RINC-0 module is versatile and not limited to LUTs alone. The approach can also be implemented in memory blocks as well, as we only have to store a table with precomputed output values for each combination of input values. Since the memory size is computed as the base-2 exponential of the number of inputs, a 30-input LUT already requires one gigabit of data. \par
     Technically, implementing a N-input LUT is efficient only for small values of N (typically under 12 inside an FPGA). In any case, it is completely unrealistic to implement a LUT for a binary circuit that has more than 40 inputs, which is still far less than the number of inputs in a typical neuron. In order to increase the number of inputs taken into account, we can build several RINC-0 DTs and combine them at higher level by applying a boosting algorithm as described in the next sub-section.
      \par

\subsubsection{RINC-1: Boosting the MAT units} \label{rinc1}

\begin{figure}[h]
    \centering
        \includegraphics[scale=0.41]{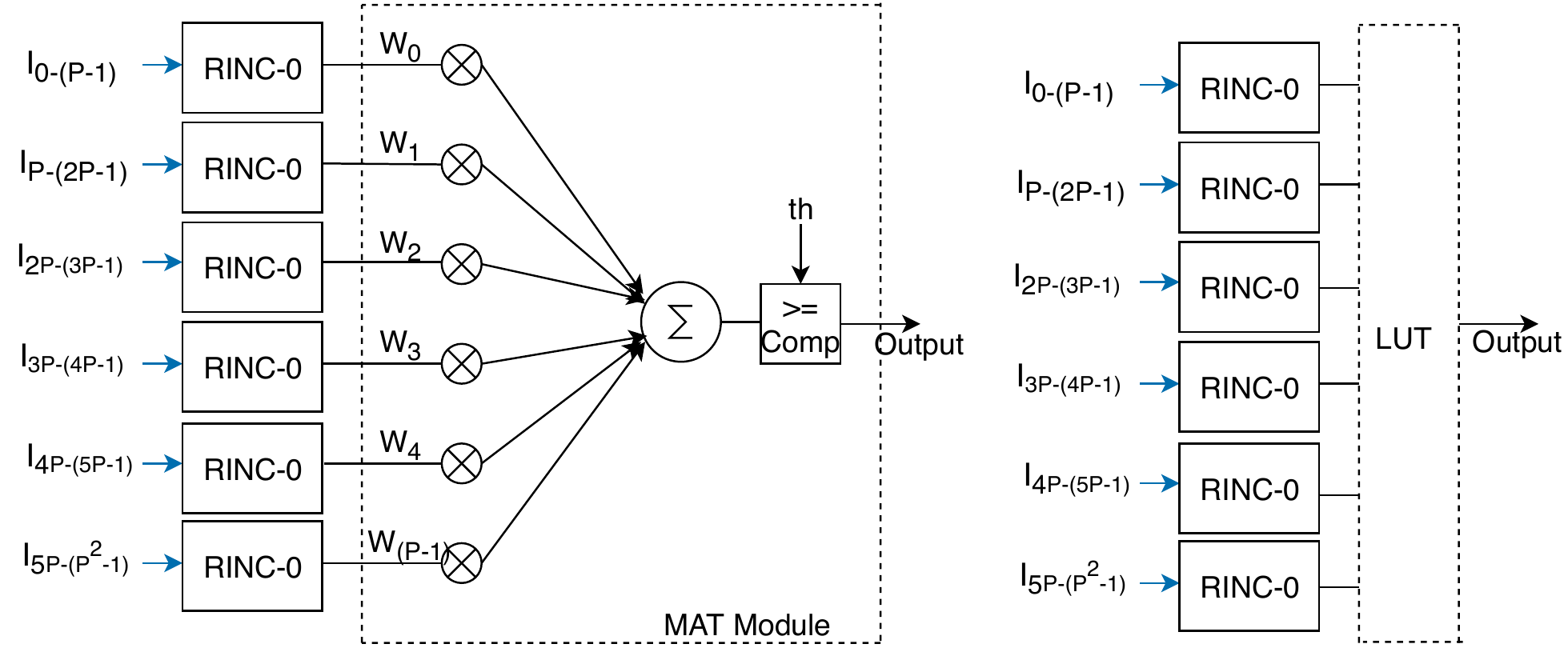}
        \caption{RINC-1 architecture with P=6}
        \label{fig36}
\end{figure}

   Even with the modified DT training approach, the RINC-0 modules have low capacity and cannot predict the output of the large binary neuron with sufficient accuracy. Hence, we increase the capacity of the weak DTs by grouping them together using a boosting algorithm. One of the most common boosting algorithms is Adaboost \cite{ratsch2001soft}, where each weak classifier (a LUT in this case) is trained sequentially and focuses on the misclassified examples of the previous classifier. Each classifier is assigned a weight ($W_{x}$), where {\it x} indicates the corresponding RINC-0 module and ranges from 0 to $P-1$. The output of each classifier is multiplied with its respective weight and added. Finally, this weighted sum is thresholded and the binary output is obtained. The architecture is detailed in the MAT unit shown in Fig. \ref{fig36}. Each MAT unit theoretically requires $P$ multiplications and $P-1$ additions. Nevertheless, since each MAT module consists of $P$ input bits and one output bit, it can also be implemented in a LUT where we pre-compute the 1-bit output for all possible $2^{P}$ inputs combinations. This MAT operation can now be performed as a single look-up operation. Thanks to the addition of the Adaboost layer, the number of inputs to the overall architecture has increased from $P$ to $P^{2}$. We denote this module as RINC-1, where-"1" is the number of Adaboost levels.\par
   However, the number of inputs to the LUT-based implementation of the MAT module is still limited. Hence, it is not possible to group more than $P$ weak classifiers. To overcome this issue, we propose a hierarchical Adaboost algorithm, which is detailed in the following sub-section.
   
\subsubsection{RINC-L: Hierarchical Adaboost Algorithm} \label{rinc2}

\begin{figure}[ht!]
    \centering
        \includegraphics[scale=0.53]{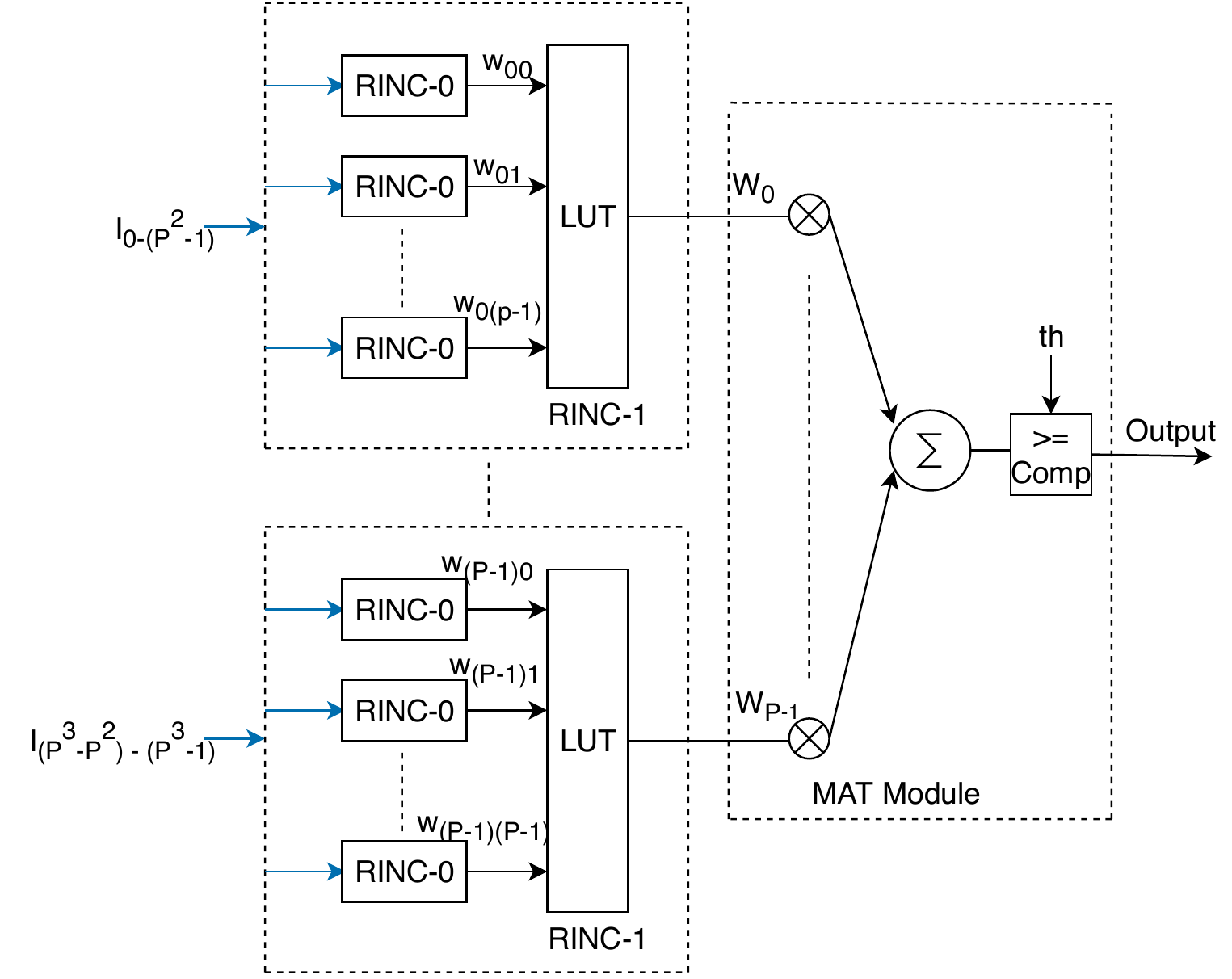}
        \caption{RINC-2 architecture}
        \label{fig_216}
        
\end{figure} 

 In FPGAs with 6-input LUTs, even with $P^{2}$ inputs, the RINC-1 module can accommodate only 36 inputs which is highly insufficient. Hence, we propose a hierarchical Adaboost algorithm to increase the capacity. Firstly, we build a RINC-1 module with $P$ DTs. We also call it a subgroup in this architecture. In each subgroup, the weights are represented as $w_{xy}$ where \textit{y} indicates the index of a RINC-0 module and \textit{x} indicates the sub-group index, as shown in Fig. \ref{fig_216}.  This subgroup is considered a weak classifier. Using the Adaboost algorithm, we construct up to $P$ subgroups. Each subgroup is assigned a weight ($W_{x}$), where x indicates the sub-group number. Hence, this creates two levels of Adaboost, one within the subgroup and second across subgroups. The binary outputs of the subgroups are multiplied, added and thresholded in a MAT module. Again this module can be implemented as a LUT. We can observe from Fig. \ref{fig_216} that adding another level of Adaboost increases the number of RINC-0 modules exponentially, thus accommodating $P^{L + 1}$ inputs. In a hierarchical Adaboost algorithm with $L$ levels and $p$ inputs per LUT, there are $P^{L}$ RINC-0 modules and $\sum_{l=0}^{L-1} P^{l}$ Look-Up based MAT modules. Thus, $$  LUTs\ required   = P^{L} + \sum_{l=0}^{L-1} P^{l} =  \sum_{l=0}^{L} P^{l} = \frac{P^{L+1}-1}{P-1}$$   Algorithm \ref{alg:example2} details the hierarchical Adaboost algorithm. We create groups of $P$ DTs together. Each DT is associated with a weight that is multiplied with the corresponding output of the DT and thresholded. Now we consider these $P$ groups of DTs as a weak classifier and assign a new weight to each group. We further build such groups of DTs and assign a weight to each of them according to the Adaboost algorithm. Again, a MAT module is required to group these sub-groups and it is implemented as a LUT. This can be viewed as 2 levels of Adaboost. Considering this RINC-2 as weak classifier, we can build sub-groups of RINC-2 classifier and group them together using a MAT module to build a RINC-3 architecture. Similarly, further levels can be added to build a RINC-L architecture. \par

   \begin{algorithm}[tb]
   \caption{RINC-L: Hierarchical Adaboost training algorithm}
   \label{alg:example2}
   \begin{algorithmic}
   \STATE {\bfseries Input:} data $X$, size $n \times F$
   \FOR{$l \rightarrow 1$ {\bfseries to} $L$}
            \STATE Construct $P$ RINC-($l-1$) classifiers
            \STATE Multiply their outputs with corresponding weights
            \STATE Sum and threshold the result
            \STATE Encode the MAT operation as LUT
            \STATE Consider these P RINC-($l-1$) classifiers as a single weak classifier
            \STATE Assign the new weights to each training example
    \ENDFOR
    \STATE {\bfseries Output:} Thresholded result of the final MAT operation

\end{algorithmic}
\end{algorithm}


\subsection{Binary to Multiclass Classification} \label{bin_mul}

The RINC-L modules can be used to implement any size of a binary neuron in the network. 
However, RINC-L being a binary classifier, it cannot be used directly for multiclass classification. Traditionally, multiclass classification using DTs has been solved using two approaches, Multiclass DTs \cite{holmes2002multiclass} and One-vs-all classifications \cite{rifkin2004defense}. Modifying the RINC architecture as Multiclass DT makes it expensive to implement in hardware. In Multiclass DTs, the leaf nodes refer to either one of the $n_{c}$ classes, where $n_{c}$ is the number of classes. This requires output of each DT to be represented over $\log_2{n_c}$ bits. Therefore, the RINC-0 and MAT modules cannot be confined to a single LUT. This effect would cascade over the entire architecture and would make it less efficient. Also, we do not consider one-vs-all classifications using Binary DTs as there is a large drop in accuracy between each individual binary classifier and the overall multi-class classifier designed by comparing the confidences \cite{abdelsalam2018polybinn}. Moreover, the one-vs-all classification requires a confidence comparison circuit that consumes more resources. \par
On the other hand, fully connected layers have been successful in multiclass classification compared to DTs. Hence, we formulate a combined approach with our RINC architecture and a fully connected layer to overcome the multiclass challenge. We preserve the output fully connected layer, while replacing the hidden layers in the classifier with our RINC-L architecture. We then adapt the output fully connected layer to work in tandem with the RINC-L architecture for multiclass classification. In a way we break the task of the classifier into two parts, first the binary classification of the hidden layer representations (\secref{r_hidden}) and then the multiclass classification of the output layer (\secref{A_output}).

 \begin{figure}[h!]
    \centering
        \includegraphics[scale=0.68]{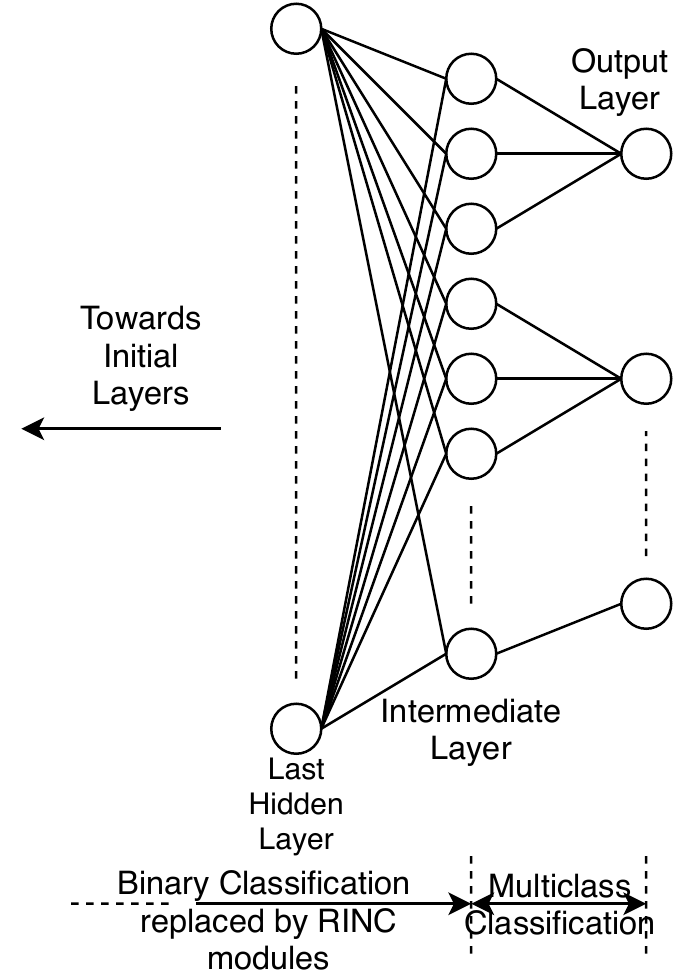}
        \caption{Intermediate layer}
        \label{fig_int}
    \end{figure} 

\subsubsection{Replacing the hidden layers} \label{r_hidden}
 We use a back to front approach where we start replacing the binary neurons in the network with our RINC-L architecture from the last layers and progressively move towards the initial layers. Given sufficient capacity of the RINC-L modules, it is possible to replace multiple layers of the neural network with a single RINC-L module. In order to minimize the number of RINC-L modules required, we add a fully connected layer with binary sigmoid activation \cite{kwan1992sigmoid} after the last hidden layer. We call this fully connected layer, intermediate layer. It consists of $n_{c} \times  P$ neurons, where $n_c$ is the number of classes in the multiclass classification. Typically, the value of $P$ ranges from 6 to 8 and the value of $n_{c}$ is 10 for the MNIST, CIFAR-10 and SVHN datasets. Hence, the intermediate layer has 60-80 neurons for these datasets. Thus, the intermediate layer with binary sigmoid activation can be viewed as a set of binary neurons. This enables us to train a RINC-L module to emulate a binary neuron representation in the intermediate layer. Similarly, binary sigmoid activation can be introduced in the earlier layers and can be replaced with RINC-L modules. Since there are fewer neurons in the intermediate layer than in the hidden layers, it takes fewer resources to train a RINC-L classifier for each of the neuron in the intermediate layer. On the other hand, this restricts the representation space. Hence, the hyper parameter $P$ must be chosen carefully to balance the trade-off between accuracy and resources. 
\subsubsection{Sparsely connected output layer}  \label{A_output}
The outputs of the RINC-L modules (emulating the intermediate layer representation) are connected to the output layer. However, this output layer needs to be optimized for LUT based implementation. Firstly, we modify the output layer to be sparsely connected to the intermediate layer. Each neuron in the output layer is connected to only $P$ neurons of the intermediate layer as shown in Fig. \ref{fig_int}. Hence, each neuron depends on $P\ $inputs and therefore can be implemented as a single Look-Up operation. Also, the output layer can be seen as a small set of $P-$input fully connected layers stacked in parallel. Since the output layer is trained as a fully connected network, it inherits all the properties of neural networks to classify multiclass data effectively. The output layer is separately retrained with RINC-L outputs to adapt the weights of the output layer. The retraining of the sparsely connected output layer for multiclass classification adapts the weights to the RINC-L binary hidden layer representations.\par
The output layer activation is not binary. Nevertheless, since it is sparsely connected to the previous layer, it can still be efficiently implemented with LUTs. Each output neuron is quantized to q-bits precision, thus requiring $q$ LUTs per output neuron. Therefore, the output layer can be implemented using $q\ \times\ n_{c}$ LUTs. This is negligible as compared to the resources required by the RINC-L modules. Thus, with the help of few more LUTs we achieve higher accuracy on multiclass classification than using costly multiclass DTs or One-Vs-All classifiers. This final architecture with multiple RINC-L modules and q-bit quantized output layer implemented as LUTs make the announced PoET-BiN.\par  

\section{Experimental Setup} \label{con3}
    In this section we detail how we train the proposed architecture and test its performance with various datasets.
    
    \begin{figure}[ht!]
    \centering
        \includegraphics[scale=0.47]{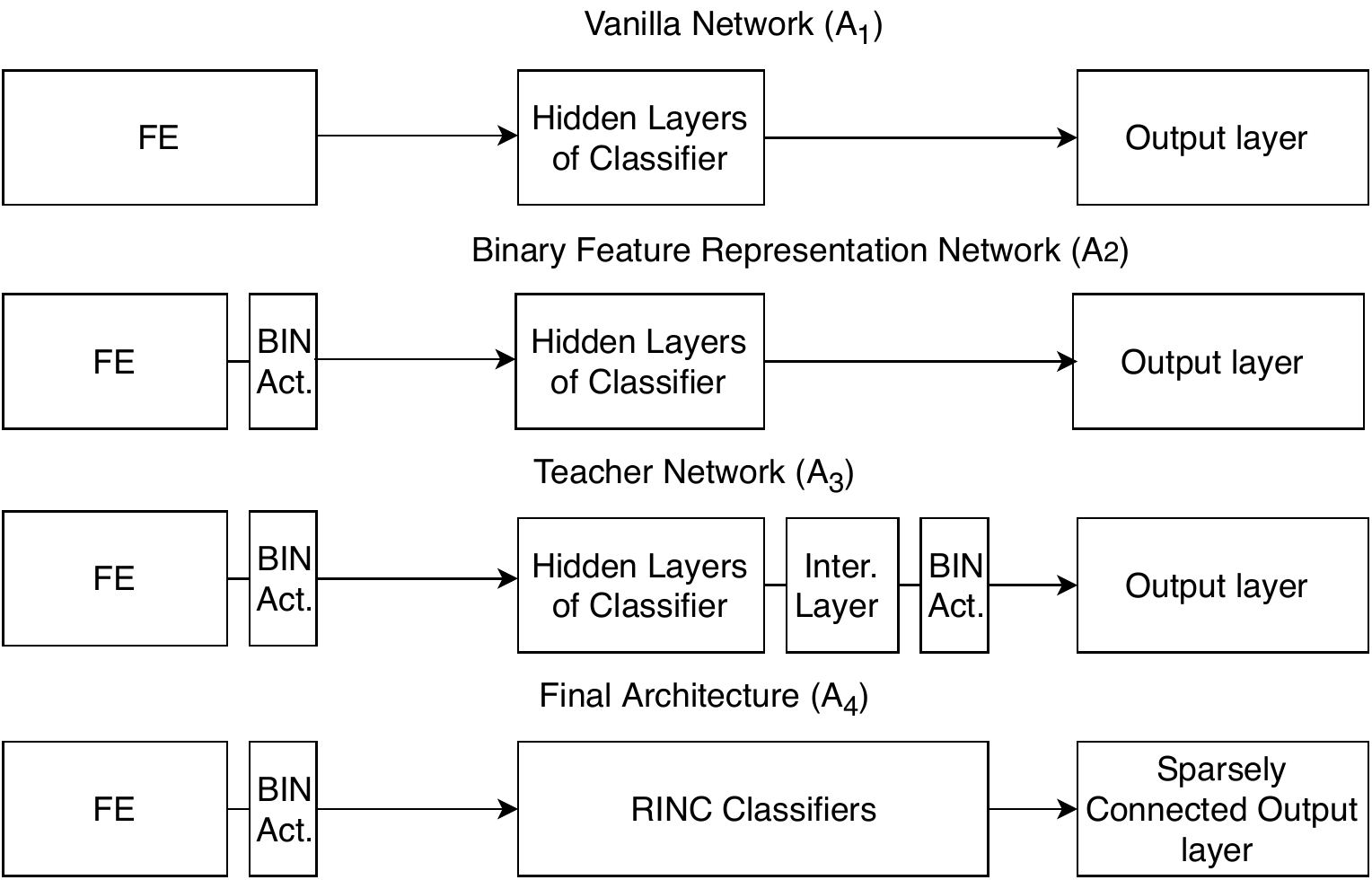}
        \caption{Overall Work Flow}
        \label{fig_workflow}
    \end{figure}

     We developed the workflow shown in Fig. \ref{fig_workflow} to train the RINC modules starting from a vanilla CNN network.
    Firstly, we use a pretrained full precision CNN as our base architecture (Vanilla network). The base architecture for each dataset is mentioned in Table \ref{tab_arch}. FE refers to the feature extractor which consists of convolutional, maxpooling and activation layers. In the vanilla network, the features are represented with full precision. However, RINC modules can only be trained on binary features. Hence, we replace the ReLU with binary sigmoid activation after the last convolutional layer to obtain the binary features. This is represented by the Bin act. module in the Binary Feature Representation Network in Fig. \ref{fig_workflow}. Further, an intermediate layer and a binary sigmoid activation are added after the last hidden layer. This forms the teacher network. Then, we replace all the hidden layers and the intermediate layer in the classifier using our RINC architecture which is the student architecture in our work. Finally, the output layer is retrained with the RINC outputs. The output layer activations are quantized to $q$ bits for efficient hardware implementation. In our test it was observed that when $q=4$, the loss in accuracy was quite significant as compared to the original floating point implementation. On the other hand, with $q=8$ the loss in accuracy was minimal. In the case when $q=16$ the accuracies were similar to that of 8-bit quantization but it requires twice the amount of LUTs as compared to 8-bit quantization. Hence we use 8-bit quantized output layer. \par 
     \par
    
    The architecture hyperparameters are listed in Table \ref{tab_arch} and explained in detail in the following sub-sections for each dataset. We use similar architecture to other recent stat-of-the-art implementations. We use techniques such as batch normalization \cite{Ioffe:2015:BNA:3045118.3045167}, exponentially decreasing learning rate, squared hinge loss \cite{rosasco2004loss} and ADAM optimizer \cite{KingmaB14} in all the vanilla networks. Also, we do not retrain with the validation set. We do not use any image augmentation techniques except for padding in CIFAR-10.\par

     \begin{table*}[h]
            \caption{Network Architecture}
            \label{tab_arch}
            \vskip 0.15in
            \begin{center}
            \begin{small}
            \begin{sc}
                \begin{tabular}{lll}
                \toprule
                Architecture (Arch.) & Symbol & Dataset\\
                \midrule
                $LeNET_{FE} - (512 FC) - (10 FC)$ & M1 & MNIST\\
                $VGG11_{FE} - (4096 FC) - (4096 FC) - (10FC)$ & C1 & CIFAR-10\\
                $VGG11_{FE} - (2048 FC) - (2048 FC)  - (10FC)$ & S1 & SVHN\\
                \bottomrule
            \end{tabular}
            \end{sc}
        \end{small}
        \end{center}
        \vskip -0.1in
    \end{table*}

      \subsection{MNIST}
    As seen in Table \ref{tab_arch}, we use the LeNet architecture for the MNIST dataset. Using two convolutional layers of $5 \times 5$ convolutions and two pooling layers of size $2 \times 2$, we transform the feature space to 512 binary features. The classifier portion consists of only one hidden fully connected layer of 512 neurons with ReLU activation and an output layer of 10 neurons. We use 8-input LUTs $(P = 8)$. Hence, the intermediate layer contains $10 \times 8 = 80$ neurons. We train a 2-level RINC (RINC-2) module with 32 DTs for each neuron in the intermediate layer. Therefore each RINC-2 module selects a maximum of 256 ( = $32 \times 8$) features from the available 512 binary features. These predicted outputs of the RINC-2 modules are used to retrain the final $8-$bit quantized output layer. \par

    \subsection{CIFAR-10}
    
     The CIFAR-10 training procedure is similar to the one used for MNIST, but for a bigger network, i.e. a VGG-11 architecture with 8 convolution layers and 3 fully connected layers. The full precision implementation proposed by Kuang \cite{kuangliu} is used as reference. The convolutional layers transform the input to a binary feature space of 512 features. There are 2 hidden fully connected layers with 4096 neurons each. To augment the capacity of RINC modules, we use $8-$input LUTs ($P=8$) and 40 DTs for each of the neuron in the intermediate layer. The intermediate layer consists of $8 \times  10 = 80$ binary neurons. The predicted outputs of the RINC-2 modules are used to retrain the output layer which is quantized to 8 bits. \par 

    \subsection{SVHN}
    SVHN is implemented with an architecture similar to the one used for CIFAR-10, except that LUTs have 6 inputs ($P=6$), leading to 36 DTs per neuron with 2 hierarchical levels (RINC-2) of Adaboost. We also use the extra dataset from the SVHN dataset for training.\par

\section{Results and Discussions} \label{res}

   We analyse the accuracy, power consumption and latency of PoET-BiN for the MNIST, CIFAR-10 and SVHN datasets. 
   \subsection{Classification Accuracy}
   
   We report four sets of accuracies for each dataset in Table \ref{ov_cl}. Firstly, we report the accuracy of the vanilla network ($A_{1}$), followed by the accuracy with binary sigmoid activation after the last convolutional layer to obtain the binary features ($A_{2}$). Then we report the accuracy after binarizing the intermediate layer ($A_{3}$). This forms the teacher network. Finally, we replace the classifier portion of the teacher network with the RINC classifiers and quantize final layer whose accuracy is reported as ($A_{4}$). This helps isolate and study the effect of each modification. We report the best accuracy achieved over different sets of hyper-parameters such as number of DTs and LUT size for RINC modules. Fig. \ref{fig_workflow} illustrates the progressive modifications with relevant accuracies.  
       
  \begin{table*}[h]
    \caption{Overall classification accuracy on MNIST, CIFAR-10 and SVHN dataset \& Comparison}
    \label{ov_cl}
    \vskip 0.15in
    \begin{center}
    \begin{small}
    \begin{sc}
    \begin{tabular}{lccccc | ccr}
    \toprule
    Arch. & Dataset & \textbf{$A_{1} (\%)$} & \textbf{$A_{2} (\%)$} & \textbf{$A_{3} (\%)$} & \textbf{$A_{4} (\%)$} & \multicolumn{3}{c}{Other works $(\%)$}\\
    & & Vanilla & & Teacher & PoET-BiN & BinaryNet & POLYBiNN & NDF \\
    \midrule
    M1 & MNIST & $99.20$ & $99.06$ & $98.93$ &  \textbf{98.15} & $98.97$ & $97.52$ & $99.42$ \\
    C1 & CIFAR-10 & $91.02$ & $89.88 $ & $89.10$  & \textbf{92.64} & $89.76$ & $91.58$ & $90.46$ \\
    S1 & SVHN & $97.36$ & $96.98$ & $96.22$  & \textbf{95.13} & $95.06$ & $94.97$ & $95.20$\\
    \bottomrule
  \end{tabular}
  \end{sc}
\end{small}
\end{center}
\vskip -0.1in
\end{table*}

 In Table \ref{ov_cl} , we observe a drop in accuracy of approximately 0.3\% for MNIST, 1.9\% for the CIFAR-10 dataset and 1.1\% for the SVHN datasets between the vanilla and teacher network ($A_1 - A_3$). This is expected as we restrict the feature space by using binary representations. This teacher network is used to train our RINC classifiers and quantized sparsely connected output layer. This results in a further dip in accuracy of $0.8\%$ for MNIST and $1\%$ for SVHN. An interesting observation in the case of CIFAR-10 is that by replacing the fully connected layers with PoET-BiN, the accuracy improves by 1.5\% for CIFAR-10. This anomaly could be due to better generalization as a result of the noise injected into the system due to the inaccuracies in intermediate layer prediction by the RINC modules. Similar observations were seen in Dropconnect \cite{wan2013regularization}.\par
 
 Now that we have obtained the final accuracy of the PoET-BiN architecture, it is necessary to have a fair comparison with other architectures present in the literature. We choose three starkly different architectures, namely BinaryNet \cite{courbariaux2016binarized}, POLYBiNN  \cite{abdelsalam2018polybinn} and Neural Decision Forest (NDF) \cite{kontschieder2015deep}. BinaryNet is a quantized CNN approach, while POLYBiNN is a complete Decision Tree approach and NDF is a hybrid mixture of both with differentiable DTs. To ensure fairness we use the same feature extractor across all architectures, and change the classifier portion of the architecture. We used our Python implementation for BinaryNet and POLYBiNN. While, Jing's Pytorch implementation of NDF \cite{jingxil} was adapted for the comparative analysis. From Table \ref{ov_cl}, we can see that our architecture performs the best in the case of CIFAR-10 and second best in case of SVHN. Though the NDF architecture performs better than PoET-BiN on MNIST and SVHN, it is not optimized for hardware implementations.\par
 For MNIST, the significant reduction in accuracy can be overcome by increasing the number of RINC classifiers. In the MNIST architecture, rather than training the RINC classifiers to predict the intermediate layer outputs, we can train a RINC classifier for each of the neuron in the only hidden layer in M1 architecture. This results in 512 RINC-2 modules. Retraining the fully connected output layer with the 512 RINC classifier outputs results in an accuracy of 98.62\% that is more closer to that of NDF. However, this implementation consumes more resources. Therefore, we do not consider this accuracy. However, it proves the versatility of the RINC architecture in implementing binary neurons. We do not implement similar architectures for SVHN and CIFAR-10, as they have significantly more neurons in the last hidden layer (2048 for SVHN and 4096 for CIFAR-10), requiring long training times. \par
Another important observation to be noted in Table \ref{ov_cl} is that the PoET-BiN architecture performs better than off-the-shelf DTs used in POLYBiNN across all datasets, in spite of them having significantly more nodes in each DT. This can be attributed to our hierarchical training algorithm and unique binary to multiclass classification technique. \par

\subsection{Power}

\begin{table}[t]
\caption{PoET-BiN power results}
\label{R_power}
\vskip 0.15in
\begin{center}
\begin{small}
\begin{sc}
\begin{tabular}{lccr}
\toprule
Power($W$) & MNIST & CIFAR-10 & SVHN \\
\midrule
Dynamic  & 0.468 & 0.300 & 0.374 \\
Static & 0.045 & 0.041 & 0.043\\
Total & 0.513 & 0.341 & 0.417\\

\bottomrule
\end{tabular}
\end{sc}
\end{small}
\end{center}
\vskip -0.1in
\end{table}

One of the most important metrics apart from accuracy for hardware implementations of neural networks is power. We compare the power consumed by the PoET-BiN architecture to the classifier portion of the vanilla neural network and quantized neural networks. We implement our PoET-BiN architecture based classifier on Spartan-6 45-nm FPGA from Xilinx to calculate the power consumption. Typically, the PoET-BiN architecture used for the three datasets consists of thousands of LUTs. It is quite cumbersome to write a HDL script for such a large implementation. Hence, we developed a Python script to generate the VHDL code automatically from the trained LUTs. The retrained final layer is also implemented on the FPGA automatically by our script. To report the power of this architecture, it has to fit in the FPGA. The Spartan-6 FPGA has 276 input/output ports but our architecture has 512 features. Hence the inputs are provided through a shift register with a single input. This enables us to fit the architecture into the target FPGA. However, this also adds some logic and signal power which is 4 mW in the case of MNIST and CIFAR-10 and 6 mW in the case of SVHN. The power consumed by the PoET-BiN architecture (subtracting the power consumed in the shift registers) are reported in Table \ref{R_power}. The outputs generated by the FPGA and those generated by PyTorch are verified in the testbench, that is automatically produced by another Python script. \par

\begin{table*}[t]
\caption{Individual operation power results}
\label{i_power}
\vskip 0.15in
\begin{center}
\begin{small}
\begin{sc}
\begin{tabular}{lcccccr}
\toprule
Operation & \multicolumn{4}{c}{Dynamic ($W$)}  & Static  & Total\\
(at 62.5 Mhz)    & \multicolumn{1}{c}{clock} &  \multicolumn{1}{c}{logic} & \multicolumn{1}{c}{signal}  & \multicolumn{1}{c}{IO} & &\\
\midrule
Multiplication (16 bits) & 0.001 & 0.001 & 0.000 & 0.020 & 0.036 & 0.058\\
Addition (16 bits) & 0.001 & 0.000 & 0.001 & 0.024 & 0.036 & 0.062\\
Multiplication (32 bits) & 0.002 & 0.001 & 0.001 & 0.035 & 0.037 & 0.076\\
Addition (32 bits) & 0.001 & 0.000 & 0.002 & 0.048 & 0.037 & 0.088\\
Multiplication (Float) & 0.005 & 0.006 & 0.005 & 0.046 & 0.037 & 0.098\\ 
Addition (Float) & 0.004 & 0.003 & 0.005 & 0.034 & 0.037 & 0.083\\

\bottomrule
\end{tabular}
\end{sc}
\end{small}
\end{center}
\vskip -0.1in
\end{table*}

\begin{table}[t]
\caption{Total mathematical operations}
\label{Tot_Op}
\vskip 0.15in
\begin{center}
\begin{small}
\begin{sc}
\begin{tabular}{lccr}
\toprule
Operation & MNIST & CIFAR-10 & SVHN \\
\midrule
Addition  & 267,264 & 18,915,328 & 5,263,360 \\
Multiplication & 267,264 & 18,915,328 & 5,263,360 \\

\bottomrule
\end{tabular}
\end{sc}
\end{small}
\end{center}
\vskip -0.1in
\end{table}

We compare the power consumed to that of the vanilla neural network and quantized neural networks. Most of the hardware implementations of neural network in the literature provide the power consumed for the entire network including the convolutional layers. It is difficult to accurately estimate the power consumed in the fully connected layers from this data. Hence we use a bottom-up approach to estimate the power of the classifier portion of these networks. Mathematical operations (multiplication and addition) and memory fetching operations consume most of the power in the fully connected layers. Power consumption of memory fetching operations depends on the evaluation platform, memory type and other factors. Hence, it is quite difficult to estimate the power required for memory fetching operations accurately without an actual implementation. On the other hand, we can estimate the power required for the mathematical operations. First, we implement a single multiplication and an addition on the same FPGA. Table \ref{i_power} provides the power consumption for a single multiplication and an addition operation. The multiplication is implemented with a Digital Signal Processor (DSP) block in the FPGA, which consumes less power compared to a LUT-based implementation. The addition operation is implemented with LUTs and dedicated carry chains. We use IP cores provided by Xilinx to implement the multiplication and addition. Now, that we know the power consumed by each operation, we calculate the total number of multiplications and additions in each of the fully connected layers of the vanilla architecture Table \ref{Tot_Op}. From these data we estimate the power consumed in the classifier portion of the vanilla and quantized neural networks except binary quantized networks. \par 
In the case of binary quantized neural network, each multiplication or addition operation consumes an insignificant amount of power. Hence, we estimate the power consumption by implementing a binary neuron. Each binary neuron consists of multiple binary multiplications (XNOR operation) followed by a tree structured adder \cite{hoe2011} and a comparator. We then multiply this value with the number of neurons in the classifier portion of the respective network for each dataset to estimate the total power consumption of classifier portion of each network. In the case of MNIST, each binary neuron consumes $34\ mW$ of logic + signal power. However, this includes the power consumed by two shift registers that are used to feed in the values for the inputs and weights. Each shift register consumes $4 mW$ of power that needs to be subtracted from the total power. Hence, the power consumed by the binary neuron portion alone is $34 - 4 - 4 = 26\ mW$. There are 522 binary neurons in the classifier portion of the M1 architecture. Therefore the total dynamic power consumed by all the binary neurons is $26 * 522 = 13.572\ W$. This value is multiplied by the time period of the clock ($16\ ns$) to obtain the energy shown in Table \ref{ov_pwr}. Similarly, we estimate the energy in the classifier portion of binary quantized networks for the other datasets.\par
Such method to estimate the power has the advantage of considering the same target device for all the estimations as illustrated in Table \ref{i_power}. Other works use the metrics proposed by Horowitz in \cite{horowitz2014} but we have not been able to find a fair estimation of a small LUT. Moreover, the power analyzer gives a detailed report on the power distribution in the FPGA. The total power can be coarsely divided into static and dynamic power. The static power, as the name suggests, is constant for a given FPGA device. The dynamic power can be further sub-divided into clock, logic, signal and IO power. The clock and IO power are also constant for a given FPGA device at a given frequency of operation. Hence, the actual energy involved in the computation of a combinational function is only concerned by the {\it logic} and {\it signal} columns of Table \ref{i_power}. Therefore, we only use these values to estimate the power of a given architecture. \par

Along with power, energy is also an important metric to be taken into consideration. The energy is calculated in Table \ref{ov_pwr} from the power values mentioned in Table \ref{R_power} and Table \ref{i_power}. To calculate the energy value we use the time period of the clock. Our PoET-BiN classifier requires single cycle to implement the inference. For the SVHN dataset, we use a RINC-2 classifier with $P=6$ that is easily implementable on Xilinx LUTs as they support 6-input LUTs. Hence, we use a 100 MHz clock for the RINC-2 classifier implementation for SVHN. On the other hand, MNIST and CIFAR-10 require RINC-2 classifiers with $P=8$. As each 8-input LUT requires four $6-$input LUTs, the critical path increases. Therefore, we use a slower 62.5 MHz clock. We can increase the frequency of the implementation by pipelining the architecture, but this will lead to more power consumption due to the extra registers. Hence, we stick to single cycle implementations. Using these information, the energy results are calculated and detailed in Table \ref{ov_pwr}. \par
\begin{table}[h!]
    \caption{Energy consumption comparison}
    \label{ov_pwr}
    \vskip 0.15in
    \begin{center}
    \begin{small}
    \begin{sc}
    \begin{tabular}{lccc}
    \toprule
    Technique  & \multicolumn{3}{c}{Energy ($J$)}  \\
    & \multicolumn{1}{c}{MNIST} &  \multicolumn{1}{c}{CIFAR-10} &  \multicolumn{1}{c}{SVHN}\\
    \midrule
    Vanilla & $8.0 \times 10^{-5} $ & $5.7 \times 10^{-3} $ & $1.6 \times 10^{-3} $\\
    1-bit Quant & $2.1 \times 10^{-7} $ & $3.9 \times 10^{-5}  $ & $9.2 \times 10^{-6} $\\
    16-bit Quant & $8.5 \times 10^{-6}$ & $6.0 \times 10^{-4}$ & $1.0 \times 10^{-4}$\\
    32-bit Quant & $1.7 \times 10^{-5}$ & $1.2 \times 10^{-3} $ & $3.6 \times 10^{-4}$\\
    PoET-BiN & $8.2 \times 10^{-9}$ & $5.4 \times 10^{-9} $ & $4.1 \times 10^{-9} $\\
    \bottomrule
  \end{tabular}
  \end{sc}
  \end{small}
  \end{center}
  \vskip -0.1in
\end{table}
We observe that the PoET-BiN architecture as compared to a full precision vanilla network consumes $1 \times 10^{4}$ times less energy in the case of MNIST, almost $1 \times 10^{6}$ for CIFAR-10 and $4 \times 10^{5}$ in the case of SVHN. Even in the case of 16-bits quantized network, the PoET-BiN architecture consumes almost $1 \times 10^{3}$ less energy in MNIST, $1 \times 10^{5}$ in CIFAR-10 and $2.5 \times 10^{4}$ in the case of SVHN. Comparing the PoET-BiN to 1-bit quantization (binary), we observe our architecture consumes $25 \times$ less energy in the case of the MNIST dataset, $7 \times 10^{3}$ less energy for the CIFAR-10 dataset and $2 \times 10^{3}$ less in the case the SVHN dataset. \par 
Actually, these values are the worst case scenario of power reduction since we do not consider the power required for memory fetching operations in vanilla and quantized neural networks. These operations are $10 \times$ more power intensive than multiplication operations \cite{horowitz2014}. On the other hand, the PoET-BiN architecture does not need any memory access operations. Hence, in reality our architecture is even more power efficient than what we report here.

\subsection{Latency and Area}

Apart from accuracy and power, embedded application are often time critical and demand low latency. Even with a single cycle implementation, the PoET-BiN architecture has a low latency. From Table \ref{latency}, we can observe that the latency is 5.85 ns in the case of SVHN, while for CIFAR-10 and MNIST it is 9.48 ns and 9.11 ns respectively. This translates to a throughput of up to $166\ M$ images per second in the case of SVHN and $100\ M$ images per second in the case of MNIST and CIFAR-10. \par

\begin{table}[h]
\caption{Implementation results of PoET-BiN}
\label{latency}
\vskip 0.15in
\begin{center}
\begin{small}
\begin{sc}
\begin{tabular}{lcccr}
\toprule
Parameters & MNIST & CIFAR-10 & SVHN \\
\midrule
Latency(ns)  & 9.11 & 9.48 & 5.85 \\
LUTs & 11899 & $9650$& $2660$\\

\bottomrule
\end{tabular}
\end{sc}
\end{small}
\end{center}
\vskip -0.1in
\end{table}

The PoET-BiN architecture stands out from other inference techniques such as quantization or pruning in this regard as our implementation is focused on making each sub operation fit a single LUT. This yields a highly optimized architecture in terms of area as seen in Table \ref{latency}. Especially in the case of SVHN, the PoET-BiN architecture with $P=6$ and 2 levels (RINC-2) requires 2660 LUTs. This can be verified with manual calculations as follows. Firstly, each RINC-0 module requires a single LUT. Then, a RINC-1 module with $P=6$, consists of 6 RINC-0 modules and a MAT module, hence requiring $6 + 1 = 7$ LUTs. A RINC-2 module consists of 6 RINC-1 module and a MAT module, thus requiring $7*6 + 1 = 43$ LUTs. Sixty such RINC-2 modules are required to emulate the intermediate layer, therefore consuming $43 * 60 = 2580$ LUTs. The final output layer consists of 10 neurons whose values are quantized to 8 bits. Hence each neuron in the output layer requires 8 LUTs. Therefore, $2580 + 80 = 2660$ LUTs are required to implement the classifier architecture for SVHN. This is the exact count given by the Xilinx synthesizer as well. As there are no overlaps between inputs in each DT, the Xilinx synthesizer cannot further simplify the design. This supports the idea that our training algorithm produces a highly efficient implementation. \par

In the case of MNIST and CIFAR-10 we use 32 and 40 DTs per RINC-2 module, respectively, with $P=8$. Since Xilinx LUTs have a maximum of 6 inputs, each $8-$input LUT requires four $6-$input Xilinx LUTs. Sometimes, this conversion results in redundancy and the synthesizer removes a few LUTs that do not affect the result. Further analysis reveals that most of the LUTs removed by the synthesizer are MAT modules. This is because some DTs have a very low weight assigned by the Adaboost algorithm, which finally does not affect the result of the MAT operation. This suggests that it is possible to further improve our training algorithm. Such opportunity is predominately visible in the case of CIFAR-10 where approximately $36\%$ of the LUTs are removed by the synthesizer producing a smaller architecture in spite of having more DTs per RINC-2 module as compared to the PoET-BiN implementation of MNIST.

\section{Related Work} \label{rel}
In the literature, there have been various implementations of DTs optimized for embedded hardware \cite{abdelsalam2018polybinn,narayanan2007interactive}. In these implementations, tackling complex classification tasks inevitably leads to deeper and more DTs. This necessitates the use of external memory to store the intermediate results. More recently, Abdelsalam et al. \cite{abdelsalam2018polybinn} implemented Binary DTs using simple AND-OR gates achieving 97 \% accuracy on the MNIST dataset. However the DTs were implemented using off-the-shelf DT libraries, leading to generic trees that were not completely optimized for the underlying LUT-based hardware. Zhou \& Feng \cite{zhou2017deep} proposed a multi-level DT architecture where the output of the DTs of the preceding layer are used as inputs to the next layer. They achieved promising results on the MNIST dataset but could only achieve 68\% on CIFAR-10. Also, there have been few attempts to combine neural networks and DTs. Kontschieder et al. first coined the term neural decision forests \cite{kontschieder2015deep}. They proposed differentiable DTs which could be trained along with convolutional layers to achieve 93.6\% top-5 accuracy on Imagenet. However, they used stochastic rounding and computations at each of the decision nodes making it unfavourable for hardware implementation.

\section{Conclusion} \label{conc}
This work introduced PoET-BiN, an architecture and its associated building algorithm, that is optimized to fit the LUTs or memory blocks in embedded systems such as FPGAs. We replaced the classifier portion of various networks with our architecture to achieve accuracies similar to the ones obtained with full precision implementations. This is because, around half of the network weights are present in the FC layers for the network models considered. We need to reduce both the computations and memory instructions to reduce the overall power of the network. Moreover, a memory fetch instruction consumes more power than a MAC operation. Hence, reducing the power in the FC layers is all the more necessary we reduce the energy consumption for the classifier portion up to six orders of magnitude compared to a floating point implementations and up to three orders of magnitude when compared to recent binary quantized neural networks. This is due to the fact that all the arithmetic operations are replaced by small LUTs on binary signals. Also, our work is quite versatile. CNNs, RNNs, LSTMs and almost every NN uses MAC operations. By binarizing certain activations, we can directly implement our RINC modules in these NNs. \par 
In future work, we will implement the convolutional layers with RINC modules and extend them to bigger datasets such as Imagenet. Another avenue of research would be to use differentiable DTs, thus training the DTs and convolutional layers together to obtain better feature representation. This may result is higher accuracies as the entire network is trained together instead of training layerwise.


\bibliography{example_paper}
\bibliographystyle{sysml2019}

%


\end{document}